\documentclass[runningheads]{llncs}

\usepackage[year=2026]{eccv}
\usepackage{eccvabbrv}
\usepackage{graphicx}
\usepackage{booktabs}
\usepackage[accsupp]{axessibility}  

\usepackage{hyperref}
\usepackage{orcidlink}
\usepackage{comment}
\usepackage{multirow}
\usepackage{multicol}
\usepackage{subcaption}
\usepackage{kotex}

\begin{document}
\title{Context-Aware Autoregressive Diffusion for Gloss-Wise Sign Language Production} 

\titlerunning{Conter paper title}

\author{JungHoon~Sung\inst{1}\textsuperscript{*} \and
 Boeun~Kim\inst{1}\textsuperscript{*} \and
 Chu~Xin\inst{1}\and
 Hyung~Jin~Chang\inst{2}\and
 ChangHo~Kim\inst{1} \and
 Sang-Il~Choi\inst{1}\textsuperscript{\dag}\and
 Younggeun~Choi\inst{1}\textsuperscript{\dag}
 }

\authorrunning{J.~Sung et al.}

\institute{
Dankook University \and
University of Birmingham
}

\maketitle

\begingroup
\renewcommand\thefootnote{}\footnotetext{
\textsuperscript{*} Equal contribution.\\
\textsuperscript{\dag} Corresponding author.
}
\endgroup

\begin{abstract}

To generate natural and accurate sentence-level sign language, synthesizing the `gloss'—the fundamental semantic unit—is essential. However, most current sign-language production (SLP) methods generate entire sequences at once. While this end-to-end approach is often efficient, it is prone to temporal drift and hand motion blur as sentences get longer, and fails to accurately control individual glosses. In this paper, we propose the \textbf{Context-aware Gloss-wise AutoRegressive Diffusion model (GARD)}, a gloss-wise diffusion framework that models coarticulation by conditioning on both semantic (linguistic) and kinematic (motion) contexts. To ensure natural continuity between gloss motions, GARD introduces two additional strategies: i) \textbf{Inter-Gloss Transition Guidance}, which applies gradient-based guidance to kinematically align inter-gloss boundaries and ensure seamless pose consistency.
ii) \textbf{Global Motion Harmonizer}, refining the entire gloss motion sequence based on the boundary poses adjusted by Inter-Gloss Transition Guidance.
Extensive experiments on Phoenix-T and CSL-Daily datasets demonstrate that GARD achieves superior performance over existing SLP methods in terms of both linguistic accuracy and motion similarity.


\end{abstract}
    
\section{Introduction}

More than 70 million Deaf individuals worldwide still face linguistic barriers in daily communication and information access\cite{deafpeople2020,deafpeople2022}. Because spoken captions alone cannot reproduce the intrinsically spatial and visual nature of signed languages\cite{SignResearch_2019, including_2021, llms_good_SL_2024}, Sign Language Production (SLP) technology has emerged as a key accessibility solution~\cite{saunders2020progressive,signdiff_fang2023,signVQNet_hwang2024,SingIDD-tang2025sign,signingAt_saunders2022signing,24neuralSignActor_CVPR,xie2024g2p, 24_t2s-gpt, 25_signs_as_token,gcdm_24tang}. SLP aims to translate spoken or written language into natural, continuous sign motions represented by 3D avatars or skeletal meshes.
\begin{figure}[t]
\centering
    \includegraphics[width=0.95\linewidth]{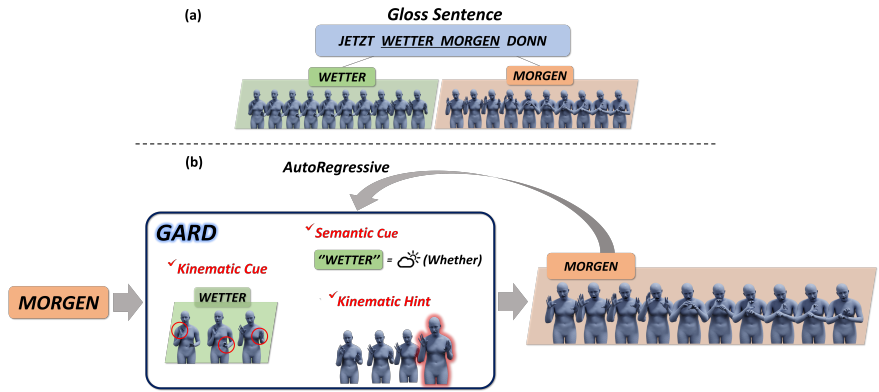}
    \caption{
Overview of sign language production process by GARD. (a) Ground-truth 3D mesh motions for two consecutive glosses. (b) Generation process of the gloss motion ``MORGEN''. In the next step, GARD will autoregressively generate ``DONN'' using semantic and kinematic cues from ``MORGEN''.
}
    \label{fig:teaser}
\vspace{-0.3cm}
\end{figure}

Most SLP models generate motion frame-by-frame conditioned on a sentence-level representation \cite{24neuralSignActor_CVPR, 24_ECCV_signavatars, 24_t2s-gpt, SingIDD-tang2025sign, xie2024g2p}. While this approach supports linguistic coherence and local temporal continuity \cite{saunders2020progressive, 24FG_signavatar, Latent_MT_xie2024sign}, it suffers from accumulated prediction errors, which progressively degrade motion quality toward the end of long sequences \cite{signVQNet_hwang2024, signdiff_fang2023, 24_t2s-gpt}. Furthermore, this approach often degrades critical semantic details, such as finger articulation and hand movements, which are essential for sign language.

To address these limitations, other studies attempted to generate full sign sentences by stitching together pre-composed motion segments corresponding to individual glosses~\cite{signingAt_saunders2022signing, 24spoken2sign_eccv}. However, such approaches remain limited because individual glosses do not explicitly encode sufficient contextual information, and the resulting motions often lack the rich variation and natural transitions observed in real signing.
This suggests that effective SLP requires modeling the gloss as the fundamental semantic unit while also explicitly addressing the transitions and continuity between consecutive glosses~\cite{sign1906/2005_stokoe}.

In sign language, \textbf{Coarticulation} between glosses plays a crucial role in shaping linguistic expression. Coarticulation refers to the phenomenon in which one gloss influences the next, producing more natural and context-dependent transitions\cite{Coarticulation_2009study,coarticulation2012,coarticulation2024}. This extends beyond simple boundary smoothness and includes broader linguistic variation, such as allomorph selection and context-dependent sign duration. In practice, the motion of the next gloss may be selected from multiple allomorphic variants depending on how the previous gloss is articulated. For example, a motion that ends with one hand must be followed by an allomorph that also begins with one hand to complete a natural flow.

Recent gloss-wise SLP methods rely on retrieving and concatenating predefined gloss units\cite{signingAt_saunders2022signing, 24spoken2sign_eccv}.
Although this strategy ensures the accurate execution of individual glosses, it remains limited in capturing the natural coarticulation present in authentic sign language.
Signing at Scale \cite{signingAt_saunders2022signing} attempted to create smooth motion by retrieving a single exemplar per gloss and applying a GAN-based blending to the transitions. However, this does not fundamentally resolve the problem, as abrupt boundary transitions, such as sudden velocity changes or mismatched rotations, can still remain.
Spoken2Sign\cite{24spoken2sign_eccv} builds a gloss dictionary and retrieves candidates, but its gloss-level retrieval process does not explicitly account for motion mismatch with the preceding gloss, often leading to limited variation and less natural transitions.

To overcome these limitations, we propose
context-aware Gloss-wise Autoregressive Diffusion model (GARD), a novel framework that generates sign motions gloss by gloss.
Fig. \ref{fig:teaser} illustrates how GARD generates a gloss sequence. GARD generates each gloss motion in sequence and autoregressively proceeds until the end of the sentence. To accurately generate the next motion, we introduce a multi-conditioning scheme that leverages semantic cues from the current and previous glosses, along with kinematic cues derived from the previous motion.

Furthermore, to maintain smooth transitions between glosses during generation, we introduce two specialized strategies within the diffusion framework using a kinematic hint (final frame of the preceding motion). 
First, Inter-Gloss Transition Guidance (IG-Guidance) aligns the boundary poses so that the first frame of the current gloss smoothly follows the final frame of the preceding gloss.
Second, Global Motion Harmonizer (GM-Harmonizer) propagates the influence of the aligned boundary frame to subsequent frames, refining the gloss motion sequence to achieve globally coherent and natural motion dynamics.

In summary, our main contributions are as follows:
\begin{itemize}
\item  We propose GARD, a novel diffusion-based framework leveraging 
both semantic (linguistic) and kinematic (motion) contexts for gloss-wise SLP.
\item We design two novel strategies for coarticulation.
IG-Guidance aligns boundary poses between consecutive gloss motions, while GM-Harmonizer propagates this alignment to refine the entire sequence for globally coherent motion.
\item We show that GARD significantly improves both linguistic accuracy and motion naturalness, outperforming state-of-the-art SLP models.
\end{itemize}

\section{Related Work}
\subsection{Autoregressive Sign-Language Generation}
Early skeleton-based research treated continuous sign production as a sequence prediction task.  
Progressive Transformer~\cite{saunders2020progressive} first mapped spoken sentences to 3D joint trajectories with an autoregressive (AR) counter-decoding schedule, establishing the PHOENIX-14T back-translation benchmark.  
Mixed SIGNal~\cite{saunders2021mixed} improved coarticulation by blending a mixture-of-motion-primitives within the same AR decoder.  
Latent Motion Transformer~\cite{Latent_MT_xie2024sign} further moves toward a tokenized regime by quantizing each pose (or video code) into discrete tokens and predicting them autoregressively, achieving higher visual fidelity without explicit pose estimation. 
By contrast, non-autoregressive approaches, such as NAT-EA \cite{huang2021towards} and Gaussian-weighted attention models \cite{non-ar_hwang2021}, aim to reduce latency. However, they often struggle with fine-grained coarticulation and rely on post-smoothing, highlighting that AR modeling remains crucial for fluent motion.

\subsection{Gloss-Wise Conditioning and Coarticulation}
A parallel line of work focuses on semantic alignment at the gloss level.
G2P-DDM~\cite{xie2024g2p} and SignVQNet~\cite{signVQNet_hwang2024} discretize gloss-conditioned poses into VQ tokens but rely on a fixed codebook that limits kinematic detail. Gloss-driven Conditional Diffusion Model (GCDM)~\cite{gcdm_24tang} injects gloss embeddings into Gaussian noise via cross-attention and aggregates multiple hypotheses. Meanwhile, Sign-IDD~\cite{SingIDD-tang2025sign} enriches each gloss pose with 4D bone vectors and an attribute-controllable diffusion module, boosting spatial precision. The discrete segment diffusion framework~\cite{discreate_tangCVPR25} treats word-level clips as inpainting masks and synthesizes transition frames, but still leaves the explicit modeling of semantic--kinematic coupling across consecutive glosses largely unaddressed.

\subsection{3D Mesh-based Sign Language Production}
Recent studies move beyond skeletons to generate full-body SMPL-X meshes.  
Stoll \etal~\cite{22_3DV_tab} synthesize temporally coherent meshes by combining back-translation loss with 2D pose guidance.  
Neural Sign Actors~\cite{24neuralSignActor_CVPR} train an anatomical graph-diffusion network on a large ASL corpus to regress photorealistic avatars.  
Dong \etal~\cite{24_word3D} leverage CLIP-conditioned diffusion for single word meshes and extend to sentence-level generation via a Transformer CVAE with curriculum learning~\cite{24FG_signavatar}.  
SignAvatars~\cite{24_ECCV_signavatars} re-target a $4D$ corpus to SMPL-X and update body, hand, and facial vertices jointly with a sentence-conditioned graph diffuser, while Spoken2Sign~\cite{24spoken2sign_eccv} constructs a gloss-indexed mesh vocabulary and interpolates retrieved poses to build a speech-to-sign pipeline.

\subsection{Diffusion-based Motion Generation}
Diffusion models have recently shown strong performance in text-driven human motion synthesis~\cite{Temos_2022}. Representative methods such as Human Motion Diffusion Model (MDM)~\cite{MDM_2022human} and MotionDiffuse~\cite{motiondiffuse_2024} show that diffusion can generate diverse and high-fidelity motion sequences. MLD~\cite{MLD_2023executing} further improves text-conditioned motion generation by performing diffusion in a latent space. Text-guided motion generation with keyframe constraints~\cite{text-guided-diff_2024} enhances controllability by enforcing sparse structural guidance.
GMD~\cite{GMD_karunratanakul2023guided} introduces trajectory-level spatial constraints to steer motion generation toward user-specified paths, while OmniControl~\cite{omnicontrol_xie2024} enables finer joint-level control over arbitrary joints and time steps during sampling. However, these methods are primarily designed for general human motion and do not explicitly model the fine-grained hand kinematics and temporal coarticulation required for sign language production.
Our GARD bridges these four research directions.  
It autoregressively denoises each gloss motion while explicitly conditioning on both (i) the semantic embeddings of the previous and current glosses and (ii) the latent kinematic state of the previous motion, thereby generating context-aware 3D sign sequences.

\section{Method}
\subsection{Overall Framework}
\begin{figure*}[t!]
\centering
    \includegraphics[width=\linewidth]{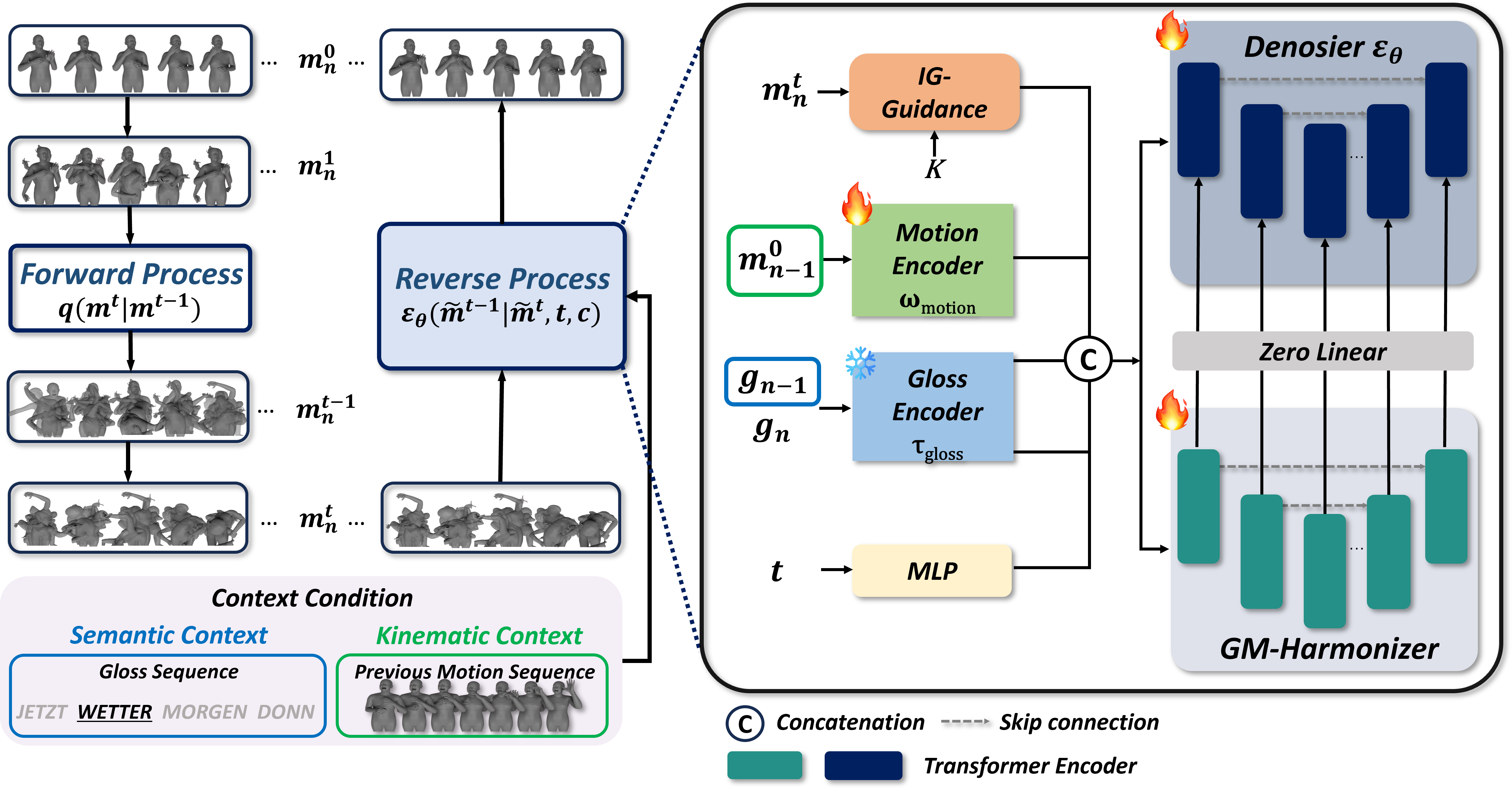}
    \caption{\textbf{Overall framework of GARD.}
    (Left) Forward and reverse diffusion processes for the $n$-th gloss.
(Right)
 During the reverse process, the denoiser $\mathbf{\epsilon}_\theta$ predicts the noise $\epsilon$ at timestep $t$, conditioned on a set of context vectors. 
 The previous gloss word $g_{n-1}$ and its corresponding motion $m_{n-1}^0$ are utilized as semantic and kinematic contexts to condition the generation of the current gloss motion. IG-Guidance refines the noised motion to reduce boundary gaps via gradient-based control, while GM-Harmonizer further enhances natural motion flow using a trainable copy of the denoiser.
}  
    \label{fig:framework}
\vspace{-0.3cm}
\end{figure*}

Our proposed framework, GARD, is illustrated in Figure \ref{fig:framework}.
GARD generates a sign language sentence from a gloss sequence, processing it gloss by gloss.
For the $n$-th gloss, the denoiser generates a clean motion $m_n^0$ from a noisy motion $m_n^t$, conditioned on the current gloss word $g_n$ and the diffusion timestep $t$.
Additionally, the previous gloss word $g_{n-1}$ and motion $m_{n-1}^0$ are encoded by the Gloss Encoder and Motion Encoder, respectively, and used as semantic and kinematic contexts.
To enhance coarticulation, we introduce Inter-Gloss Transition Guidance (IG-Guidance) and Global Motion Harmonizer (GM-Harmonizer).
The IG-Guidance refines $m_n^t$ via gradient-based control, yielding $\check{m}_n^t$ whose denoised motion has its first frame adjusted to minimize the boundary gap with the previous gloss.
The GM-Harmonizer, a structural replica of the denoiser, further refines the motion using a residual training strategy, promoting a natural flow from the first frame.

\subsection{Data Representation}
Let a gloss sequence be ${G}=(g_1, \dots, g_N)$ and its corresponding 3D mesh motion sequence be ${M}=(m_1, \dots, m_N)$ parameterized by the SMPL-X body model\cite{smplx_2019}, where $N$ is the number of glosses.
Each motion sequence $m_n \in \mathbb{R}^{F \times D}$ contains $F$ frames, and every frame is represented by a gloss motion feature of dimension $D$.
We extract motion features using a feature-extraction pipeline \cite{Guo_2022_CVPR, amass2020} widely adopted in motion diffusion research.
Our motion feature is constructed from the positions and rotations of the upper-body and hand joints, as they are crucial for capturing sign language expression.
Specifically, for each motion $m_n$, $r_p$ denotes the 3D root joint position in global coordinates. $j_p$ represents the relative coordinates of each joint, converted from global joint coordinates. $j_r$ corresponds to the joint rotations encoded as 6D representations\cite{6d_repre_2019}.
The overall motion feature sequence is defined as:
\begin{equation}
m_n = \bigl[r_p,\;j_p,\;j_r\bigr]^{1:F}.
\end{equation}

\subsection{Context-aware Denoiser}
\textbf{Forward process.}
GARD is a diffusion-based generator with an autoregressive design that sequentially produces motion for each gloss in a sentence.
During training, the motion feature at each gloss level is progressively corrupted by adding Gaussian noise $\epsilon\sim\mathcal{N}(0,\mathbf I)$ at every diffusion step $t= 1,\dots,T$, following a predefined noise schedule.
\begin{equation}
q\;\!\bigl(\mathbf{m}_n^{t}\mid\mathbf{m}_n^{t-1}\bigr)
=\mathcal{N}\;\!\bigl(
\sqrt{\alpha^t }\,\mathbf{m}_n^{t-1},\;
\!\bigl(1-\alpha^t\bigr)\mathbf I \;
\bigr).
\end{equation}
In the denoising phase, the denoiser receives the noisy sample and directly predicts the injected noise at each frame, progressively removing it over $T$ steps to reconstruct the clean motion $\tilde{m}_{n}^{0}$.

\noindent\textbf{Reverse process.}
Our denoiser is built upon a Transformer Encoder \cite{transformer_2017,MDM_2022human, MLD_2023executing}, an architecture widely adopted for motion and sequence data over the U-Net denoisers \cite{Unet-2015} common in image diffusion. We further integrate a skip-connection structure \cite{23_longskip, text-guided-diff_2024} that fuses low-level and high-level representations, enabling the preservation of subtle local joint movements and the maintenance of visual continuity. This combined design allows the model to generate fine-grained, natural sign language motions with smoother transitions. The denoiser jointly captures correlations among all input feature vectors and directly predicts the injected noise $\epsilon_\theta(\tilde{m}_{n}^{t})$ at the positions corresponding to the noisy frames in its output.
We use a pretrained Semantic Context Extractor, $\tau_{gloss}(\cdot)$, which extracts semantic context vector from the glosses.
The context vector of previous gloss $\tau_{gloss}(g_{n-1}) \in \mathbb{R}^{1 \times 1024}$ and the current gloss  $\tau_{gloss}(g_{n}) \in \mathbb{R}^{1 \times 1024}$ are employed as conditions of the diffusion model.
The Semantic Context Extractor is obtained by finetuning an mBART model that was pretrained on large-scale multilingual data \cite{mbart2020} as many recent SLT/SLP pipelines have adopted mBART and demonstrated its effectiveness for semantic encoding \cite{24spoken2sign_eccv, TwoStream_chen2022two, SMMT_2022CVPR, cosign_2023ICCV, toward2024ENNLP}.
Additionally, the motion $m_{n-1}$ corresponding to the previous gloss $g_{n-1}$ is encoded as a kinematic feature using a trainable Kinematic Context Extractor, implemented as a simple Transformer encoder\cite{transformer_2017}, $\omega_{\mathrm{motion}}(m_{n-1}) \in \mathbb{R}^{1 \times d}$.  
To enable feature fusion, the semantic context vectors are first projected to the same dimensionality as the kinematic context vector, i.e.,
$\tilde{\tau}_{gloss}(g)=\mathrm{Proj}\bigl(\tau_{gloss}(g)\bigr)$, where $\tilde{\tau}_{gloss}(g)\in\mathbb{R}^{1 \times d}$.
We then concatenate the semantic context and the kinematic context to construct a context condition vector as
\begin{equation}
\mathbf{c}_n
= \bigl[
    \;\tilde\tau_{gloss}(g_n)
    \; ;
    \; \tilde\tau_{gloss}(g_{n-1})
    \; ; 
    \;\omega_{motion} (m_{n-1})  
  \bigr].
  \label{equ:condition}
\end{equation}

Subsequently, $\mathbf{c}_n$ is entered into the denoiser together with the noisy motion vector at time step $t$, denoted as $\tilde{m}_{n}^{t}$:
The self-attention modules within the denoiser learn the correlations among the input features, enabling the noisy motion vector $\tilde{m}_{n}^{t}$ to incorporate both contextual and motion information effectively.
During training, the ground truth motion ${m}_{n-1}^{0}$ is used as input through teacher forcing \cite{teacher_forcing_2015}, whereas during inference, the previously generated motion $\tilde{m}_{n-1}^{0}$ is used in an autoregressive manner.
For diffusion training, we define the noise reconstruction objective \cite{ddpm_2020} as the mean squared error between the target noise $\epsilon$ and the predicted noise $\varepsilon_\theta\!\bigl(\tilde{m}_{n},c_n,t\bigr)$:
 
\begin{equation}
\mathcal{L}_{\text{diff}}
:= \mathbb{E}_{\epsilon,t}\Bigl[ 
   \bigl\|
      {\epsilon} -\varepsilon_\theta\!\bigl(\tilde{m}_{n},c_n,t\bigr)
  \bigr\|_2^{2}
   \Bigr].
\end{equation}

During sampling, we compute $\tilde{m}_n^{t-1}$ from $m_n^t$ according to the reverse schedule rule given in Eq. (\ref{equ:reverse}).
Repeating this procedure $T$ steps yields the fully restored motion $\hat{m}_n^{0}$.
Consequently, the model learns to perform reverse denoising while maintaining linguistic continuity across the sentence:
\begin{align}
\tilde{m}_n^{t-1}
&= \frac{1}{\sqrt{\alpha^t}}
  \left(
     \tilde{m}_n^t
     - (1 - \alpha^t)\,\varepsilon_\theta\bigl(\tilde{m}_n^t,c_n,t\bigr)
  \right)
  + \sigma^t\,z, \label{equ:reverse1} \\
\alpha^t &= 1 - \beta^t, \qquad
(\sigma^t)^2 = \beta^t\,\frac{1-\alpha^{t-1}}{1-\alpha^t},
\label{equ:reverse}
\end{align}
where $\beta^t$ denotes the variance of the Gaussian noise injected at step $t$, while $\alpha^t$ is the complementary scaling factor. The term $\sigma^t$ represents the standard deviation of the Gaussian noise re-injected at reverse-diffusion step $t$ according to the noise schedule.

\noindent\textbf{Classifier Free Guidance.}
We apply the classifier-free guidance \cite{cfg2022classifier} to balance the diversity of generated sign language motions with fidelity to the context conditions. To enable the denoiser $\varepsilon_\theta$ to learn both conditioned and unconditioned noise distributions during training, we randomly drop the conditioning $c_n$ for 10\% of the training samples. For inference, given a guidance scale $s$, the guided denoiser is defined as:
\begin{equation}
\begin{split}
\varepsilon_\theta^{s}\!\bigl(\tilde{m}_{n},c_n,t\bigr)
= (1-s)\cdot\,\varepsilon_\theta\!\bigl(\tilde{m}_{n},\emptyset,t\bigr)
+ s\cdot\,\varepsilon_\theta\!\bigl(\tilde{m}_{n},c_n,t\bigr).
\end{split}
\label{eq:scaled_noise}
\end{equation}

Increasing $s$ places greater emphasis on the conditioning $c_n$, resulting in motions that more closely align with the intended context but exhibit reduced diversity. Conversely, smaller $s$ promote greater diversity in the generated motions at the expense of strict conformity to the context.

\subsection{Inter-Gloss Transition Guidance}
Despite conditioning on the previous motion context, the pose gap at the boundary between consecutive generated glosses is not fully eliminated.
To address this issue, we introduce IG-Guidance to ensure a natural transition between the last pose of the previous gloss $g_{n-1}$ (the kinematic hint $K_{\text{hint}}$) and the first pose of the current gloss $g_n$. Inspired by gradient-based motion control techniques~\cite{GMD_karunratanakul2023guided,omnicontrol_xie2024},
IG-Guidance refines the predicted first-frame pose by minimizing the rotational discrepancy between the predicted rotations and those of the kinematic hint.

IG-Guidance computes a gradient to pull the joint rotation matrices of the current gloss's first frame ($R_{tgt}$) closer to the joint rotation matrices of the Kinematic Hint ($R_{K}$). We utilize the Geodesic distance $D_{\mathrm{geo}}$~\cite{2020_geodesic}. While $L_2$ measures the straight-line Euclidean distance, $3$D rotations lie on the rotation manifold $SO(3)$. Therefore, the meaningful notion of discrepancy is the Geodesic distance, i.e., the shortest path along the manifold, rather than the Euclidean straight-line distance. 
During a single diffusion step $t$, the gradient is applied for $i$ iterative refinement steps to progressively reduce the gap. The guidance strength is controlled by a scale parameter $\tau$. A single refinement step is defined as:
\begin{align}
&\check{m}_n^t = m_n^t  - \tau \nabla_{m_n^t}  D_{\mathrm{geo}}(R_{tgt}, R_{K}), \\
D_{\mathrm{geo}}(&R_{tgt}, R_{K})
= \arccos\!\left(\frac{\operatorname{trace}\!\big(R_{tgt}^\top R_{K}\big) - 1}{2}\right).
\end{align}

This gradient-based update encourages the initial pose of the current gloss to geometrically align with the final pose of the previous gloss, leading to temporally smooth and visually coherent sign motion generation.

\begin{figure}[t]

\centering
    \includegraphics[height=0.4\linewidth]{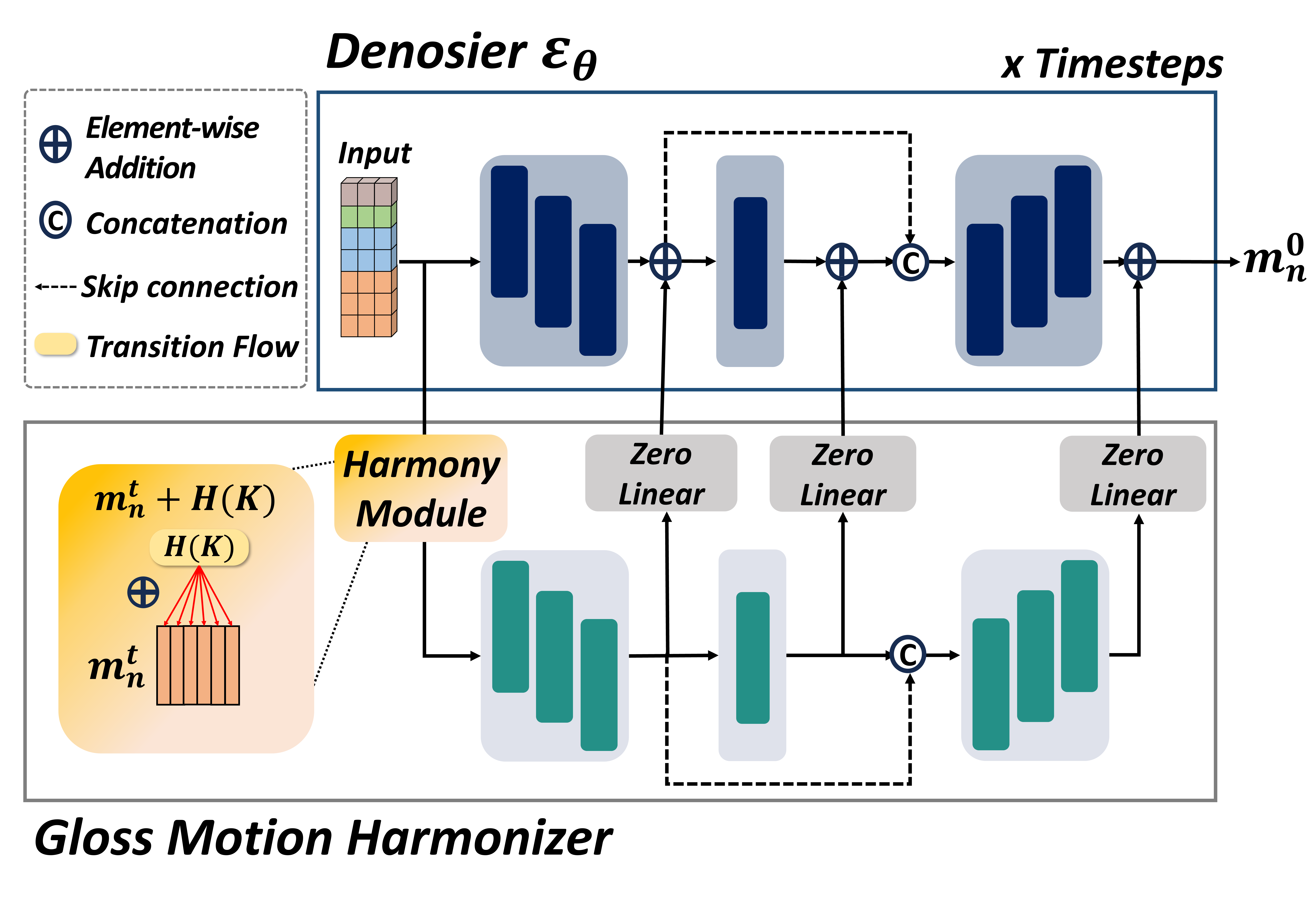}
    \caption{
   Illustration of Our GARD's Gloss Motion Harmonizer.  
    }
    \label{fig:gmh}
\vspace{-0.3cm}
\end{figure}

\subsection{Global Motion Harmonizer}
Using IG-Guidance, the first frame is generated to smoothly connect with the previous gloss. GM-Harmonizer then propagates the influence of this frame through the subsequent frames. OmniControl~\cite{omnicontrol_xie2024} guides other joints to move organically along a single determined joint trajectory using a ControlNet-based residual learning strategy\cite{23_controlnet}.
Inspired by this residual learning approach, we propose GM-Harmonizer to refine the pose sequence of the subsequent frames while preserving the already adjusted first frame, ensuring a natural and continuous motion transition. 

As illustrated in Figure \ref{fig:gmh}, GM-Harmonizer is implemented as a trainable copy of the main denoiser $\mathbf{\varepsilon}_\theta$. The GM-Harmonizer receives the same context conditions $c_n$ and noisy motion $m_n^t$ as the denoiser, along with the kinematic hint $K$ obtained from IG-Guidance.
As shown in Figure \ref{fig:gmh}, the kinematic hint $K \in \mathbb{R}^{1\times m}$ is encoded through the Harmony module $H(\cdot)$ and added to the noisy motion via element-wise addition with temporal broadcasting. $\mathrm{Broadcast}_L(\cdot)$ replicates the $1\times d$ vector along the temporal dimension to match the $L\times d$ shape. 
\begin{equation}
\begin{split}
{}^{\text{\scriptsize gmh}}\mathbf{m}^{n}_{t}
&= \mathbf{m}^{n}_{t} + \mathrm{Broadcast}_{L}\!\left(H(K)\right), \\
H(K)
&= \mathrm{GELU}\!\left(\mathrm{LayerNorm}\!\left(\mathrm{Linear}(K)\right)\right).
\end{split}
\label{eq:gmh}
\end{equation}

These noisy motion ${}^{\text{\scriptsize gmh}}\mathbf{m}^{n}_{t}$, together with the context conditions are fed into the GM-Harmonizer's Transformer Encoder layers, where they self-attend to effectively learn the temporal motion flow. The hierarchical intermediate features produced by the GM-Harmonizer are integrated into the Denoiser via zero-initialized linear layers. 
These zero-linear layers act as gates, allowing the original denoiser to generate motion without any influence from the GM-Harmonizer at the beginning of training. As training progresses, corrections derived from the GM-Harmonizer are gradually injected into the denoiser. This correction is applied hierarchically and implicitly across each layer block.
\section{Experiments}
\textbf{Dataset.}
We evaluate our framework on two large-scale continuous sign language benchmarks: PHOENIX-2014T \cite{PHOENIX-T_camgoz2018} and CSL-Daily \cite{2021_csl_daily}.
PHOENIX-2014T is a widely used German Sign Language dataset containing 7,096 training, 5,196 development, and 6,420 test samples with 1,066 unique glosses. Despite its limited vocabulary, it poses strong challenges for generalization due to diverse sentence compositions. CSL-Daily is a larger Chinese Sign Language corpus comprising over 20,000 sentences from 50 signers and about 2,000 glosses.


\begin{table*}[t]
\centering
\scriptsize
\setlength{\tabcolsep}{1.5pt}
\renewcommand{\arraystretch}{1.12}
\caption{Comparison with state-of-the-art methods on Phoenix-T and CSL-Daily.\\
(K: keypoint-based, M: mesh-based; Gls: gloss-dependent; B4: BLEU-4, R: ROUGE, D-P-J: DTW-PA-JPE, D-J: DTW-JPE; $\ast$: re-implemented under a fair setting following the original paper.)}
\label{tab:table_phoenix_csl}

\begin{tabular}{@{}p{2.0cm}lcccccccccccccc@{}}
\toprule
\multirow{4}{*}{Methods} & \multirow{4}{*}{Cat.} & \multirow{4}{*}{Gls.} &
\multicolumn{6}{c}{Phoenix-T} &
\multicolumn{6}{c}{CSL-Daily} \\
\cmidrule(lr){4-9}\cmidrule(lr){10-15}

& & & \multirow{2}{*}{B4$\uparrow$} & \multirow{2}{*}{R$\uparrow$} &
\multicolumn{2}{c}{D-P-J$\downarrow$} &
\multicolumn{2}{c}{D-J$\downarrow$} &
\multirow{2}{*}{B4$\uparrow$} & \multirow{2}{*}{R$\uparrow$} &
\multicolumn{2}{c}{D-P-J$\downarrow$} &
\multicolumn{2}{c}{D-J$\downarrow$} \\

\cmidrule(lr){6-7}\cmidrule(lr){8-9}
\cmidrule(lr){12-13}\cmidrule(lr){14-15}
& & & & & Body & Hand & Body & Hand
& & & Body & Hand & Body & Hand \\
\midrule
PT* \cite{saunders2020progressive} & K & $\checkmark$
& 4.94 & - & 13.67 & 11.95 & 15.01 & 31.77
& 3.07 & - & 15.98 & 12.91 & 16.30 & 32.63 \\

NAT-EA \cite{huang2021towards} & K & $\checkmark$
& 6.66 & 19.43 & - & - & - & -
& - & - & - & - & - & - \\

DET \cite{2022_DET} & K & $\checkmark$
& 5.32 & 17.85 & - & - & - & -
& - & - & - & - & - & - \\

GenOBT \cite{22_tang_genobt} & K & $\checkmark$
& 8.01 & 23.49 & - & - & - & -
& - & - & - & - & - & - \\

D3DPsign \cite{23_shan_d3dp} & K & $\checkmark$
& 5.25 & 17.55 & - & - & - & -
& - & - & - & - & - & - \\

G2P-DDM* \cite{xie2024g2p} & K & $\checkmark$
& 7.13 & 20.11 & - & - & - & -
& 6.52 & 20.54 & - & - & - & - \\

GCDM \cite{gcdm_24tang} & K & $\checkmark$
& 7.91 & 23.20 & - & - & - & -
& - & - & - & - & - & - \\

Sing-IDD* \cite{SingIDD-tang2025sign} & K & $\checkmark$
& 9.08 & 26.58 & - & - & - & -
& 8.13 & 27.43 & - & - & - & - \\

\midrule
T2M \cite{22_3DV_text2mesh} & M & 
& 5.81 & - & 13.48 & 12.06 & 14.04 & 31.64
& 5.11 & - & 13.47 & 12.10 & 13.76 & 30.37 \\

T2S-GPT \cite{24_t2s-gpt} & M & 
& 9.06 & - & 10.38 & 6.47 & 11.65 & 19.09
& 8.94 & - & 11.94 & 5.93 & 12.32 & 15.43 \\ 

M-GPT \cite{23_motiongpt} & M & 
& 9.68 & - & 9.45 & 3.41 & 10.42 & 9.08
& 8.84 & - & 10.81 & 3.78 & 11.58 & 11.31 \\

SOKE* \cite{25_signs_as_token} & M & 
& 11.84 & - & \textbf{4.72} & 1.39 & \textbf{6.12} & 7.80
& 11.21 & - & \textbf{6.32} & 1.83 & \textbf{7.41} & 9.79 \\

\midrule
\textbf{GARD} & M & $\checkmark$
& \textbf{13.16} & \textbf{37.47} & 5.03 & \textbf{1.21} & 6.69 & \textbf{6.90}
& \textbf{12.73} & \textbf{36.76} & 6.98 & \textbf{1.56} & 7.73 & \textbf{9.32} \\

\bottomrule
\end{tabular}%
\end{table*}

\begin{figure*}[t!]
\centering
    \includegraphics[width=0.95\linewidth]{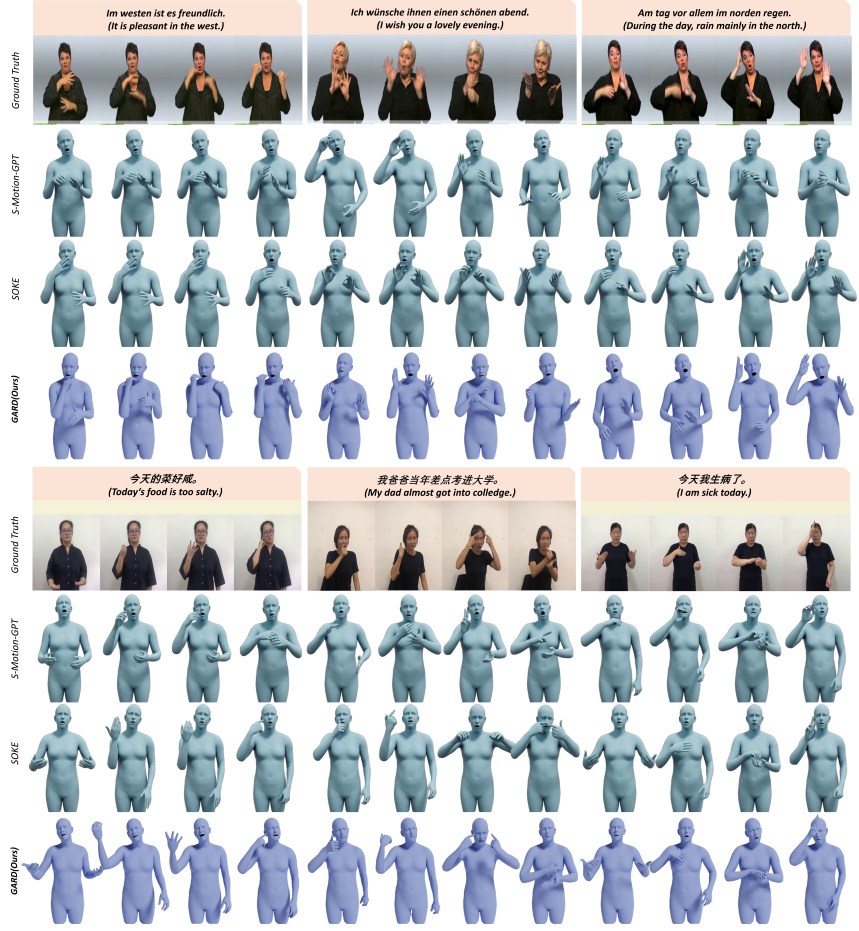}
    \caption{
    Visual comparisons with SOTA Method, SOKE\cite{25_signs_as_token} and S-MotionGPT \cite{23_motiongpt}}
    \label{fig:qualitative_result}
\vspace{-0.3cm}
\end{figure*}
\noindent\textbf{Implementation.}
For sign motion, we utilize the SMPL-X 3D mesh data curated in previous research~\cite{25_signs_as_token} for our training.
To segment sentence-level videos into gloss units, we use a pretrained sign language recognition model~\cite{TwoStream_chen2022two} for automatic gloss boundary detection. 
During inference, we employ a pretrained length predictor module to estimate the length of the gloss to be generated. This predictor takes the previous motion sequence and the embeddings of both the previous and current glosses as input. The motion sequence is contextually encoded via a Transformer encoder, while the two gloss embeddings act as a query to perform attention pooling over this encoded sequence. The summarized vector extracted through this process is then used to directly predict the length of the current gloss.
Our denoiser performs 1,000 diffusion steps with a batch size of 64. It comprises 4 attention heads and 9 layers, with a model dimension of 512. The denoiser is optimized using AdamW \cite{adamW_2017decoupled} with a learning rate of $10^{-4}$. The guidance scale $s$ is set to 7. Training is conducted for a total of 1,000 epochs.
The Kinematic motion encoder is implemented as a Transformer encoder with 4 attention heads and two layers. 

\noindent\textbf{Evaluation Metrics.}
To comprehensively evaluate our proposed model, we assess the generated sign language sequences from two primary perspectives: linguistic accuracy and motion quality. Following prior works~\cite{saunders2020progressive, 25_signs_as_token, SingIDD-tang2025sign, 24neuralSignActor_CVPR}, we employ a pretrained back-translator, which converts the generated 3D joint sequences back into spoken (textual) language.  
The back-translator is trained on 3D skeletal data and enables indirect evaluation of linguistic fidelity. We report the BLEU \cite{bleu2002} and ROUGE\cite{rouge2004} scores computed from the back-translated sentences, which measure how accurately the generated sign sequences convey the intended meaning in textual form.
To assess the motion kinematic similarity between the generated motion and the ground-truth, we use Dynamic Time Warping (DTW) \cite{dtw_origin_1994}, following the method described in \cite{25_signs_as_token,ham2pose_dtw, 24neuralSignActor_CVPR}. DTW-JPE measures the average joint position error between aligned frames, while DTW-PA-JPE evaluates the joint error after Procrustes alignment, thus focusing on pose similarity independent of global translation and scale.

\subsection{Comparison with State-of-the-Art Methods}
\noindent\textbf{Quantitative Results.}
In Table~1, we present a comprehensive comparison of our proposed GARD model with several recent methods on the test sets of Phoenix-T and CSL-Daily. We compare performance across two main categories of sign language generation approaches, namely keypoint-based and $3$D mesh-based methods. In addition, the ``Gls.'' column indicates whether each method is gloss-dependent, i.e., whether it explicitly relies on gloss information during generation.
The proposed method consistently performs strongly on both Phoenix-T and CSL-Daily, achieving state-of-the-art (SOTA) results in BLEU, ROUGE, and DTW-Hand. In sign language, hand position and finger shape are the most critical components for conveying semantic meaning. Therefore, the DTW-Hand metric is directly correlated with the model's linguistic performance. Our improved DTW-Hand scores indicate that GARD more accurately captures the fine-grained hand position and movement essential to sign generation. This enhanced hand-motion fidelity directly leads to higher BLEU and ROUGE scores confirming that our model achieves both visual naturalness and superior linguistic accuracy.

\noindent\textbf{Qualitative Results.}
We perform a qualitative comparison against recent 3D mesh-based studies. As can be seen in Figure~\ref{fig:qualitative_result},
our GARD model accurately captures the Ground-Truth (GT) sign features, demonstrating advantages over SOTA models, particularly in the detailed movement of the fingers and the overall magnitude of the sign motion.
For instance, comparing the CSL-Daily results, we observe distinct differences in fine-grained motions, such as the clenching of the fist or the extension of the fingers. SOKE\cite{25_signs_as_token} encodes decoupled motions using a VQ-VAE-based tokenization scheme. Because VQ-VAE maps diverse poses to a single token, it loses subtle details and produces averaged poses during decoding. This convergence to neutral states with smaller motion magnitude explains its lower DTW-Body score. S-MotionGPT\cite{23_motiongpt} also showed poor sign motion accuracy in our visual comparison. Ultimately, GARD demonstrates superior performance in both quantitative metrics and visual quality.

\begin{table*}[t]
\centering
\begin{minipage}[t]{0.48\textwidth}
    \centering
    \captionof{table}{Ablation study of Context Condition settings (B1 : BLEU-1 , B4 : BLEU-4 , R : ROUGE)}
    \label{tab:ablation_condition}
    \vspace{0.3cm}
    \scriptsize
\setlength{\tabcolsep}{2pt}
\renewcommand{\arraystretch}{1.1}
\begin{tabular}{lcccc}
\toprule
\multirow{2}{*}{\textbf{Variants}} &
\multirow{2}{*}{B1$\uparrow$} & \multirow{2}{*}{B4$\uparrow$} &
\multirow{2}{*}{R$\uparrow$} & \multirow{2}{*}{WER$\downarrow$} \\
& & & & \\
\midrule
Base               & 23.57 & 8.83 & 30.21 & 76.24\\
Base + Kinematic   & 25.13 & 9.34 & 30.28 & 75.49\\
Base + Semantic    & 26.22 & 9.87 & 32.02 & 73.88\\
\midrule
\textbf{GARD}      & \textbf{28.38} & \textbf{10.63} & \textbf{33.76} & \textbf{72.71}\\
\bottomrule
\end{tabular}
\end{minipage}
\hfill
\begin{minipage}[t]{0.48\textwidth}
    \centering
    \captionof{table}{Ablation study of the IG-Guidance (IGG) and GM-Harmonizer (GMH)  (B4 : BLEU-4 , R : ROUGE) }
    \label{tab:ablation_coar}
    \vspace{0.3cm}
    \scriptsize
\setlength{\tabcolsep}{2pt}
\renewcommand{\arraystretch}{1.1}
\begin{tabular}{lccccc}
\toprule
\multirow{2}{*}{\textbf{Variants}} &
\multirow{2}{*}{B4$\uparrow$} &
\multirow{2}{*}{R$\uparrow$} &
\multirow{2}{*}{WER$\downarrow$} &
\multicolumn{2}{c}{DTW-JPE$\downarrow$} \\
\cmidrule(lr){5-6}
& & & & Body & Hand \\
\midrule
Base           & 10.63 & 33.76 & 72.71 & 9.71 & 9.30\\
Base + IGG     & 10.88 & 34.04 & 73.13 & 9.42 & 9.05\\
Base + GMH     & 13.11 & 36.90 & 65.38 & 8.81 & 8.03\\
\midrule
\textbf{GARD}  & \textbf{13.16} & \textbf{37.47} & \textbf{63.24} & \textbf{6.69} & \textbf{6.90}\\
\bottomrule
\end{tabular}
\end{minipage}

\end{table*}
\subsection{Ablation Study}
\noindent\textbf{Effects of Context Conditions.} 
Table~\ref{tab:ablation_condition} shows the effects of different context conditions on our model. We perform comparative experiments using a ``Base'' method that generates motion solely based on the current gloss word. The ``Base + Semantic Context'' variant generates motion by combining the preceding gloss word. In contrast, The ``Base + Kinematic Context'' variant utilizes the motion of the previous gloss.
Incorporating the semantic context leads to a significant performance improvement across all metrics.
This is particularly evident in the Word Error Rate (WER), where lower scores are better. Adding Semantic Context alone provides the most substantial error reduction, a much larger impact than adding only Kinematic Context.
The combination of both contexts achieves the greatest overall gain, reaching the highest BLEU/ROUGE scores and the lowest WER.
This suggests that both types of representations (semantic and kinematic) are critical factors in gloss-level SLP.

\begin{table}[t]
\centering
\scriptsize
\setlength{\tabcolsep}{3pt}
\renewcommand{\arraystretch}{1.1}
\caption{Ablation Study for Denoiser on Phoenix-T and CSL-Daily datasets. The first row shows our default full model configuration (9 layers, w/ skip, Self-attention, GMH, Geodesic), and the subsequent rows show the performance when specific components are altered.}
\label{tab:ablation_settings}
\begin{tabular}{@{}l cccc c cccc c@{}}
\toprule
\multirow{3}{*}{Settings} & \multicolumn{5}{c}{Phoenix-T} & \multicolumn{5}{c}{CSL-Daily} \\
\cmidrule(lr){2-6}\cmidrule(lr){7-11}

& \multirow{2}{*}{B4$\uparrow$} & \multirow{2}{*}{R$\uparrow$} & \multirow{2}{*}{WER$\downarrow$} & \multicolumn{2}{c}{DTW-JPE$\downarrow$} 
& \multirow{2}{*}{B4$\uparrow$} & \multirow{2}{*}{R$\uparrow$} & \multirow{2}{*}{WER$\downarrow$} & \multicolumn{2}{c}{DTW-JPE$\downarrow$} \\

\cmidrule(lr){5-6} \cmidrule(lr){10-11}
& & & & Body & Hand & & & & Body & Hand \\
\midrule
\textbf{Full Model} & \textbf{13.16} & \textbf{37.47} & \textbf{63.42} & \textbf{6.69} & \textbf{6.90} 
& \textbf{12.73} & \textbf{36.76} & 61.03 & 7.73 & \textbf{9.32} \\
\midrule
7 layers (vs. 9)
& 12.10 & 35.83 & 65.43 & 8.02 & 8.53 
& 11.33 & 34.32 & 64.10 & 10.06 & 11.47 \\
11 layers (vs. 9)
& 12.42 & 36.75 & 66.89 & 8.11 & 7.88 
& 12.31 & 36.54 & \textbf{60.72} & \textbf{7.56} & 9.78 \\

w/o skip  (vs. w/ skip)
& 10.15 & 33.06 & 72.26 & 9.27 & 9.18 
& 9.72 & 31.16 & 71.12 & 11.29 & 12.45 \\

Cross-att (vs. Self-att)
& 11.13 & 33.82 & 71.53 & 8.92 & 9.53 
& 10.64 & 32.24 & 72.33 & 11.04 & 11.76 \\

Residual (vs. GMH)
& 8.31 & 25.20 & 77.38 & 14.84 & 18.36 
& 7.54 & 23.59 & 80.12 & 19.93 & 17.60 \\

L2 Dist (vs. Geodesic) 
& 12.42 & 36.28 & 64.14 & 8.24 & 7.87 
& 12.20 & 35.95 & 62.33 & 8.56 & 10.19 \\
\bottomrule
\end{tabular}
\end{table}
\noindent\textbf{Effects of Denoiser Architecture.}
We conduct an ablation study on several architecture design choices, as summarized in Table~\ref{tab:ablation_settings}. First, we evaluate the effect of the denoiser depth on both datasets. On Phoenix-T, the 9-layer model achieves the best performance across all metrics. On CSL-Daily, while 11 layers yield marginally better WER and DTW-Body scores, 9 layers still deliver superior translation metrics (BLEU-4, ROUGE) and hand kinematics (DTW-Hand). Based on this trade-off, we adopt 9 layers for the final model. We further find that removing skip connections leads to a substantial performance drop, indicating that skip connections play an important role in preserving intermediate information for detailed sign motion generation. We also observe that Self-attention consistently outperforms cross-attention, proving its superior capability in modeling mutual interactions among context conditions. 

\noindent\textbf{Ablation on the Proposed Modules.}
We further analyze the design choices within our proposed modules. Specifically, we compare GMH with a Simple Residual variant, which naively injects the last frame of the previous gloss motion into the denoiser through a direct residual connection. The clear performance gain demonstrates that GMH is substantially more effective at handling motion transitions. 
In addition, for IG-Guidance, replacing the standard $L_2$ distance with the Geodesic distance further improves performance, leading to more accurate and expressive pose sequence generation.

\begin{figure}[t]
\centering
    \includegraphics[width=1\linewidth]{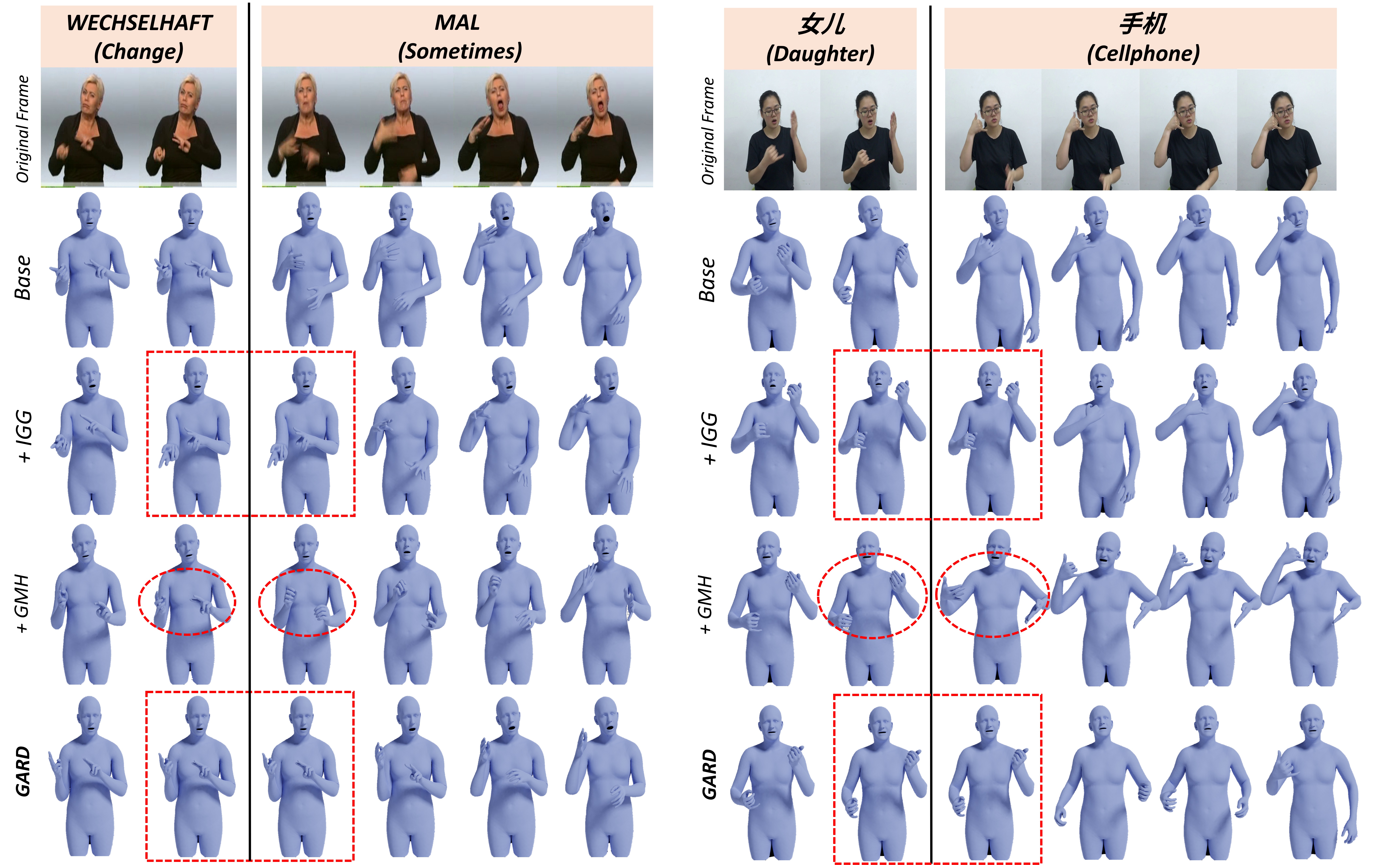}
    \vspace{-0.2cm}
    \caption{
    Qualitative ablation results for IG-Guidance and GM-Harmonizer
    }
    \vspace{-0.3cm}
    \label{fig:qualitative_result_2}
\vspace{-0.3cm}
\end{figure}

\noindent\textbf{Effects of IG-Guidance and GM-Harmonizer.} 
We conduct an ablation study to verify the contribution of each key component of coarticulation: IG-Guidance and GM-Harmonizer. 
The Base model uses all semantic and kinematic context inputs but excludes both IG-Guidance and GM-Harmonizer. As shown in Table~\ref{tab:ablation_coar}, GM-Harmonizer significantly improves performance in both DTW and BLEU compared to the baseline. Moreover, the combination of both components leads to the greatest performance improvement. Figure ~\ref{fig:qualitative_result_2} provides qualitative results that present the effects of each component by visualizing the generated motion at the boundary between two consecutive glosses. While the base model successfully generates the basic motion for each gloss, it fails to produce a proper transition between consecutive glosses. 
In contrast, the model with GM-Harmonizer applied begins the next sign motion from a state (e.g., hand shape and body position) that is noticeably closer to the previous gloss. This confirms that it improves temporal smoothness. However, it still shows differences in fine-grained details and fails to align the spatial gap completely. 
In the case of adding IG-Guidance, the first frame of the generated sign motion was well-aligned with the previous motion, but we observe that the gap increases sharply at the second frame.
As a result, our full model effectively addresses these issues. IG-Guidance enforces precise spatial alignment at the boundary, while the GM-Harmonizer ensures a smooth temporal flow originates from that aligned pose. As a result, the full model produces the most natural and seamless coarticulated motion among the compared variants.
\section{Conclusion}
We propose a gloss-wise autoregressive diffusion approach for sign language production. By combining both semantic and kinematic context, our model generates accurate motion while capturing context-dependent variation. In addition, we introduce two modules IG-Guidance and the GM-Harmonizer to ensure natural transitions between consecutive glosses. Comprehensive evaluations on benchmark datasets demonstrate the superior performance of our model.

\noindent\textbf{Limitations.} GARD relies on both a gloss-by-gloss autoregressive approach and an iterative diffusion sampling process; its inference speed remains a challenge for real-time interaction.

\bibliographystyle{splncs04}
\bibliography{main}
\end{document}